# Provably Bounded-Optimal Agents

**Stuart J. Russell**                                    RUSSELL@CS.BERKELEY.EDU
*Computer Science Division, University of California*
*Berkeley, CA 94720, USA*

**Devika Subramanian**                                   DEVIKA@CS.CORNELL.EDU
*Computer Science Department, Cornell University*
*Ithaca, NY 14853, USA*

## Abstract

Since its inception, artificial intelligence has relied upon a theoretical foundation centred around *perfect rationality* as the desired property of intelligent systems. We argue, as others have done, that this foundation is inadequate because it imposes fundamentally unsatisfiable requirements. As a result, there has arisen a wide gap between theory and practice in AI, hindering progress in the field. We propose instead a property called *bounded optimality*. Roughly speaking, an agent is bounded-optimal if its program is a solution to the constrained optimization problem presented by its architecture and the task environment. We show how to construct agents with this property for a simple class of machine architectures in a broad class of real-time environments. We illustrate these results using a simple model of an automated mail sorting facility. We also define a weaker property, *asymptotic bounded optimality* (ABO), that generalizes the notion of optimality in classical complexity theory. We then construct *universal* ABO programs, i.e., programs that are ABO no matter what real-time constraints are applied. Universal ABO programs can be used as building blocks for more complex systems. We conclude with a discussion of the prospects for bounded optimality as a theoretical basis for AI, and relate it to similar trends in philosophy, economics, and game theory.

## 1. Introduction

Since before the beginning of artificial intelligence, philosophers, control theorists and economists have looked for a satisfactory definition of rational behaviour. This is needed to underpin theories of ethics, inductive learning, reasoning, optimal control, decision-making, and economic modelling. Doyle (1983) has proposed that AI itself be defined as the computational study of rational behaviour—effectively equating rational behaviour with intelligence. The role of such definitions in AI is to ensure that theory and practice are correctly aligned. If we define some property $P$, then we hope to be able to design a system that provably possesses property $P$. Theory meets practice when our systems exhibit $P$ in reality. Furthermore, that they exhibit $P$ in reality should be something that we actually care about. In a sense, the choice of what $P$ to study determines the nature of the field.

There are a number of possible choices for $P$:

- *Perfect rationality*: the classical notion of rationality in economics and philosophy. A perfectly rational agent acts at every instant in such a way as to maximize its expected utility, given the information it has acquired from the environment. Since action selection requires computation, and computation takes time, perfectly rational agents do not exist for non-trivial environments.

- *Calculative rationality*: the notion of rationality studied in AI. A calculatively rational agent eventually returns what would have been the rational choice at the beginning of its deliberation. There exist systems such as influence diagram evaluators that exhibit this property for a decision-theoretic definition of rational choice, and systems such as nonlinear planners that exhibit it for a logical definition of rational choice. This is assumed to be an interesting property for a system to exhibit since it constitutes an "in-principle" capacity to do the right thing. Calculative rationality is of limited value in practice, because the actual behaviour exhibited by such systems is absurdly far from being rational; for example, a calculatively rational chess program will choose the right move, but may take $10^{50}$ times too long to do so. As a result, AI system-builders often ignore theoretical developments, being forced to rely on trial-and-error engineering to achieve their goals. Even in simple domains such as chess, there is little theory for designing and analysing high-performance programs.

- *Metalevel rationality*: a natural response to the problems of calculative rationality. A metalevel rational system optimizes over the object-level computations to be performed in the service of selecting actions. In other words, for each decision it finds the optimal combination of computation-sequence-plus-action, under the constraint that the action must be selected by the computation. Full metalevel rationality is seldom useful because the metalevel computations themselves take time, and the metalevel decision problem is often more difficult than the object-level problem. Simple approximations to metalevel rationality have proved useful in practice—for example, metalevel policies that limit lookahead in chess programs—but these engineering expedients merely serve to illustrate the lack of a theoretical basis for agent design.

- *Bounded optimality*: a bounded optimal agent behaves as well as possible given its computational resources. Bounded optimality specifies optimal *programs* rather than optimal *actions* or optimal *computation sequences*. Only by the former approach can we avoid placing constraints on intelligent agents that cannot be met by any program. Actions and computations are, after all, generated by programs, and it is over programs that designers have control.

We make three claims:

1. A system that exhibits bounded optimality is desirable in reality.

2. It is possible to construct provably bounded optimal programs.

3. Artificial intelligence can be usefully characterized as the study of bounded optimality, particularly in the context of complex task environments and reasonably powerful computing devices.

The first claim is unlikely to be controversial. This paper supports the second claim in detail. The third claim may, or may not, stand the test of time.

We begin in section 2 with a necessarily brief discussion of the relationship between bounded optimality and earlier notions of rationality. We note in particular that some important distinctions can be missed without precise definitions of terms. Thus in section 3 we provide formal definitions of agents, their programs, their behaviour and their rationality.

Together with formal descriptions of task environments, these elements allow us to prove that a given agent exhibits bounded optimality. Section 4 examines a class of agent architectures for which the problem of generating bounded optimal configurations is efficiently soluble. The solution involves a class of interesting and practically relevant optimization problems that do not appear to have been addressed in the scheduling literature. We illustrate the results by showing how the throughput of an automated mail-sorting facility might be improved. Section 5 initiates a discussion of how bounded optimal configurations might be learned from experience in an environment. In section 6, we define a weaker property, *asymptotic bounded optimality* (ABO), that may be more robust and tractable than the strict version of bounded optimality. In particular, we can construct *universal* ABO programs. A program is universally ABO if it is ABO regardless of the specific form of time dependence of the utility function.[1] Universal ABO programs can therefore be used as building blocks for more complex systems. We conclude with an assessment of the prospects for further development of this approach to artificial intelligence.

## 2. Historical Perspective

The classical idea of perfect rationality, which developed from Aristotle's theories of ethics, work by Arnauld and others on choice under uncertainty, and Mill's utilitarianism, was put on a formal footing in *decision theory* by Ramsey (1931) and vonNeumann and Morgenstern (1947). It stipulates that a rational agent always act so as to maximize its *expected* utility. The expectation is taken according to the agent's own beliefs; thus, perfect rationality does not require omniscience.

In artificial intelligence, the logical definition of rationality, known in philosophy as the "practical syllogism", was put forward by McCarthy (1958), and reiterated strongly by Newell (1981). Under this definition, an agent should take any action that it believes is guaranteed to achieve any of its goals. If AI can be said to have had a theoretical foundation, then this definition of rationality has provided it. McCarthy believed, probably correctly, that in the early stages of the field it was important to concentrate on "epistemological adequacy" before "heuristic adequacy" — that is, capability in principle rather than in practice. The methodology that has resulted involves designing programs that exhibit calculative rationality, and then using various speedup techniques and approximations in the hope of getting as close as possible to perfect rationality. Our belief, albeit unproven, is that the simple agent designs that fulfill the specification of calculative rationality may not provide good starting points from which to approach bounded optimality. Moreover, a theoretical foundation based on calculative rationality cannot provide the necessary guidance in the search.

It is not clear that AI would have embarked on the quest for calculative rationality had it not been operating in the halcyon days before formal intractability results were discovered. One response to the spectre of complexity has been to rule it out of bounds. Levesque and Brachman (1987) suggest limiting the complexity of the environment so that calculative and perfect rationality coincide. Doyle and Patil (1991) argue strongly against this position.

---

1. This usage of the term "universal" derives from its use in the scheduling of randomized algorithms by Luby, Sinclair and Zuckerman (1993).

Economists have used perfect rationality as an abstract model of economic entities, for the purposes of economic forecasting and designing market mechanisms. This makes it possible to prove theorems about the properties of markets in equilibrium. Unfortunately, as Simon (1982) pointed out, real economic entities have limited time and limited powers of deliberation. He proposed the study of *bounded rationality*, investigating "... the shape of a system in which effectiveness in computation is one of the most important weapons of survival." Simon's work focussed mainly on *satisficing* designs, which deliberate until reaching some solution satisfying a preset "aspiration level." The results have descriptive value for modelling various actual entities and policies, but no general prescriptive framework for bounded rationality was developed. Although it proved possible to calculate optimal aspiration levels for certain problems, no structural variation was allowed in the agent design.

In the theory of games, bounds on the complexity of players have become a topic of intense interest. For example, it is a troubling fact that defection is the only equilibrium strategy for unbounded agents playing a fixed number of rounds of the Prisoners' Dilemma game. Neyman's theorem (Neyman, 1985), recently proved by Papadimitriou and Yannakakis (1994), shows that an essentially cooperative equilibrium exists if each agent is a finite automaton with a number of states that is less than exponential in the number of rounds. This is essentially a bounded optimality result, where the bound is on space rather than on speed of computation. This type of result is made possible by a shift from the problem of selecting *actions* to the problem of selecting *programs*.

I. J. Good (1971) distinguished between perfect or "type I" rationality, and metalevel or "type II" rationality. He defines this as "the maximization of expected utility *taking into account deliberation costs*." Simon (1976) also says: "The global optimization problem is to find the least-cost or best-return decision, *net* of computational costs." Although type II rationality seems to be a step in the right direction, it is not entirely clear whether it can be made precise in a way that respects the desirable intuition that computation is important. We will try one interpretation, although there may be others.[2] The key issue is the space over which the "maximization" or "optimization" occurs. Both Good and Simon seem to be referring to the space of possible deliberations associated with a *particular decision*. Conceptually, there is an "object-level machine" that executes a sequence of computations under the control of a "meta-level machine." The outcome of the sequence is the selection of an external action. An agent exhibits type II rationality if at the end of its deliberation and subsequent action, its utility is maximized compared to all possible deliberate/act pairs in which it could have engaged. For example, Good discusses one possible application of type II rationality in chess programs. In this case, the object-level steps are node expansions in the game tree, followed by backing up of leaf node evaluations to show the best move. For simplicity we will assume a per-move time limit. Then a type II rational agent will execute whichever sequence of node expansions chooses the best move, of all those that finish before the time limit.[3] Unfortunately, the computations required in the "metalevel machine" to select the object-level deliberation may be extremely expensive. Good actually proposes a fairly simple (and nearly practical) metalevel decision procedure for chess, but it is far from optimal. It is hard to see how a type II rational agent could justify executing a suboptimal object-level computation sequence if we limit the scope of the optimization problem to a single decision. The difficulty can only be resolved by thinking about the design of the agent program, which generates an unbounded set of possible deliberations in response to an unbounded set of circumstances that may arise during the life of the agent.

---

2. For example, it is conceivable that Good and Simon really intended to refer to finding an agent design that minimizes deliberation costs in general. All their discussions, however, seem to be couched in terms of finding the right deliberation for each decision. Thus, type II or metalevel rationality coincides with bounded optimality if the bounded optimal agent is being designed for a *single* decision in a *single* situation.

Philosophy has also seen a gradual evolution in the definition of rationality. There has been a shift from consideration of *act utilitarianism* — the rationality of individual acts — to *rule utilitarianism*, or the rationality of general policies for acting. This shift has been caused by difficulties with individual versus societal rationality, rather than any consideration of the difficulty of computing rational acts. Some consideration has been given more recently to the tractability of general moral policies, with a view to making them understandable and usable by persons of average intelligence (Brandt, 1953). Cherniak (1986) has suggested a definition of "minimal rationality", specifying lower bounds on the reasoning powers of any rational agent, instead of upper bounds. A philosophical proposal generally consistent with the notion of bounded optimality can be found in Dennett's "Moral First Aid Manual" (1986). Dennett explicitly discusses the idea of reaching equilibrium within the space of decision procedures. He uses as an example the PhD admissions procedure of a philosophy department. He concludes, as do we, that the best procedure may be neither elegant nor illuminating. The existence of such a procedure, and the process of reaching it, are the main points of interest.

Many researchers in AI, some of whose work is discussed below, have worked on the problem of designing agents with limited computational resources. The 1989 AAAI Symposium on AI and Limited Rationality (Fehling & Russell, 1989) contains an interesting variety of work on the topic. Much of this work is concerned with metalevel rationality.

Metareasoning — reasoning about reasoning — is an important technique in this area, since it enables an agent to control its deliberations according to their costs and benefits. Combined with the idea of *anytime* (Dean & Boddy, 1988) or *flexible* algorithms (Horvitz, 1987), that return better results as time goes by, a simple form of metareasoning allows an agent to behave well in a real-time environment. A simple example is provided by iterative-deepening algorithms used in game-playing. Breese and Fehling (1990) apply similar ideas to controlling multiple decision procedures. Russell and Wefald (1989) give a general method for precompiling certain aspects of metareasoning so that a system can efficiently estimate the effects of individual computations on its intentions, giving fine-grained control of reasoning. These techniques can all be seen as approximating metalevel rationality; they provide useful insights into the general problem of control of reasoning, but there is no reason to suppose that the approximations used are optimal in any sense.

The intuitive notion of bounded optimality seems to have become current in the AI community in the mid-1980's. Horvitz (1987) uses the term *bounded optimality* to refer to "the optimization of computational utility given a set of assumptions about expected problems and constraints in reasoning resources." Russell and Wefald (1991) say that an agent exhibits bounded optimality for a given task environment "if its program is a solution to the constrained optimization problem presented by its architecture." Recent work by Etzioni (1989) and Russell and Zilberstein (1991) can be seen as optimizing over a well-defined set of agent designs, thereby making the notion of bounded optimality more precise. In the next section, we build a suitable set of general definitions from the ground up, so that we can begin to demonstrate examples of provably bounded optimal agents.

---

3. One would imagine that in most cases the move selected will be the same move selected by a Type I agent, but this is in a sense "accidental" because further deliberation might cause the program to abandon it.

## 3. Agents, Architectures and Programs

Intuitively, an *agent* is just a physical entity that we wish to view in terms of its *perceptions* and *actions*. What counts in the first instance is what it does, not necessarily what it thinks, or even whether it thinks at all. This initial refusal to consider further constraints on the internal workings of the agent (such as that it should reason logically, for example) helps in three ways: first, it allows us to view such "cognitive faculties" as planning and reasoning as occurring *in the service of* finding the right thing to do; second, it makes room for those among us (Agre & Chapman, 1987; Brooks, 1986) who take the position that systems can do the right thing without such cognitive faculties; third, it allows more freedom to consider various specifications, boundaries and interconnections of subsystems.

We begin by defining agents and environments in terms of the actions and percepts that they exchange, and the sequence of states they go through. The agent is described by an *agent function* from percept sequences to actions. This treatment is fairly standard (see, e.g., Genesereth & Nilsson, 1987). We then go "inside" the agent to look at the *agent program* that generates its actions, and define the "implementation" relationship between a program and the corresponding agent function. We consider performance measures on agents, and the problem of designing agents to optimize the performance measure.

### 3.1 Specifying agents and environments

An agent can be described abstractly as a mapping (the *agent function*) from percept sequences to actions. Let $\mathbf{O}$ be the set of percepts that the agent can receive at any instant, and $\mathbf{A}$ be the set of possible actions the agent can carry out in the external world. Since we are interested in the behaviour of the agent over time, we introduce a set of time points or instants, $\mathbf{T}$. The set $\mathbf{T}$ is totally ordered by the $<$ relation with a unique least element. Without loss of generality, we let $\mathbf{T}$ be the set of non-negative integers.

The percept history of an agent is a sequence of percepts indexed by time. We define the set of percept histories to be $\mathbf{O^T} = \{O^\mathbf{T} : \mathbf{T} \to \mathbf{O}\}$. The prefix of a history $O^\mathbf{T} \in \mathbf{O^T}$ till time $t$ is denoted $O^t$ and is the projection of $O^\mathbf{T}$ on $[0..t]$. We can define the set of percept history prefixes as $\mathbf{O}^t = \{O^t \mid t \in \mathbf{T}, O^\mathbf{T} \in \mathbf{O^T}\}$. Similarly, we define the set of action histories $\mathbf{A^T} = \{A^\mathbf{T} : \mathbf{T} \to \mathbf{A}\}$. The set of action history prefixes is $\mathbf{A}^t$, defined as the set of projections $A^t$ of histories $A^\mathbf{T} \in \mathbf{A^T}$.

**Definition 1** *Agent function: a mapping*

$$f : \mathbf{O}^t \to \mathbf{A}$$

*where*

$$A^{\mathbf{T}}(t) = f(O^t)$$

Note that the agent function is an entirely abstract entity, unlike the agent program that implements it. Note also that the "output" of the agent function for a given percept sequence may be a null action, for example if the agent is still thinking about what to do. The agent function specifies what the agent does at each time step. This is crucial to the distinction between perfect rationality and calculative rationality.

Agents live in environments. The states of an environment $E$ are drawn from a set $\mathbf{X}$. The set of possible state trajectories is defined as $\mathbf{X^T} = \{X^{\mathbf{T}} : \mathbf{T} \to \mathbf{X}\}$. The agent does not necessarily have full access to the current state $X^{\mathbf{T}}(t)$, but the percept received by the agent does depend on the current state through the *perceptual filtering function* $f_p$. The effects of the agent's actions are represented by the environment's *transition function* $f_e$, which specifies the next state given the current state and the agent's action. An environment is therefore defined as follows:

**Definition 2** *Environment $E$: a set of states $\mathbf{X}$ with a distinguished initial state $X_0$, a transition function $f_e$ and a perceptual filter function $f_p$ such that*

$$
\begin{aligned}
X^{\mathbf{T}}(0) &= X_0 \\
X^{\mathbf{T}}(t+1) &= f_e(A^{\mathbf{T}}(t), X^{\mathbf{T}}(t)) \\
O^{\mathbf{T}}(t) &= f_p(X^{\mathbf{T}}(t))
\end{aligned}
$$

The state history $X^{\mathbf{T}}$ is thus determined by the environment and the agent function. We use the notation *effects*$(f, E)$ to denote the state history generated by an agent function $f$ operating in an environment $E$. We will also use the notation $[E, A^t]$ to denote the state history generated by applying the action sequence $A^t$ starting in the initial state of environment $E$.

Notice that the environment is discrete and deterministic in this formulation. We can extend the definitions to cover non-deterministic and continuous environments, but at the cost of additional complexity in the exposition. None of our results depend in a significant way on discreteness or determinism.

## 3.2 Specifying agent implementations

We will consider a physical agent as consisting of an architecture and a program. The architecture is responsible for interfacing between the program and the environment, and for running the program itself. With each architecture $M$, we associate a finite programming language $\mathcal{L}_M$, which is just the set of all programs runnable by the architecture. An *agent program* is a program $l \in \mathcal{L}_M$ that takes a percept as input and has an internal state drawn from a set $\mathbf{I}$ with initial state $i_0$. (The initial internal state depends on the program $l$, but we will usually suppress this argument.) The set of possible internal state histories is $\mathbf{I^T} = \{I^{\mathbf{T}} : \mathbf{T} \to \mathbf{I}\}$. The prefix of an internal state history $I^{\mathbf{T}} \in \mathbf{I^T}$ till time $t$ is denoted $I^t$ and is the projection of $I^{\mathbf{T}}$ on $[0..t]$.

**Definition 3** *An architecture $M$ is a fixed interpreter for an agent program that runs the program for a single time step, updating its internal state and generating an action:*

$$M : \mathcal{L}_M \times \mathbf{I} \times \mathbf{O} \to \mathbf{I} \times \mathbf{A}$$

*where*

$$\langle I^{\mathbf{T}}(t+1), A^{\mathbf{T}}(t) \rangle = M(l, I^{\mathbf{T}}(t), O^{\mathbf{T}}(t))$$

Thus, the architecture generates a stream of actions according to the dictates of the program. Because of the physical properties of the architecture, running the program for a single time step results in the execution of only a finite number of instructions. The program may often fail to reach a "decision" in that time step, and as a result the action produced by the architecture may be null (or the same as the previous action, depending on the program design).

### 3.3 Relating agent specifications and implementations

We can now relate agent programs to the corresponding agent functions. We will say that an agent program $l$ running on a machine $M$ *implements* the agent function $Agent(l, M)$. The agent function is constructed in the following definition by specifying the action sequences produced by $l$ running on $M$ for all possible percept sequences. Note the importance of the "Markovian" construction using the internal state of the agent to ensure that actions can only be based on the past, not the future.

**Definition 4** *A program $l$ running on $M$ implements the agent function $f = Agent(l, M)$, defined as follows. For any environment $E = (\mathbf{X}, f_e, f_p)$, $f(O^t) = A^{\mathbf{T}}(t)$ where*

$$
\begin{aligned}
\langle I^{\mathbf{T}}(t+1), A^{\mathbf{T}}(t) \rangle &= M(l, I^{\mathbf{T}}(t), O^{\mathbf{T}}(t)) \\
O^{\mathbf{T}}(t) &= f_p(X^{\mathbf{T}}(t)) \\
X^{\mathbf{T}}(t+1) &= f_e(A^{\mathbf{T}}(t), X^{\mathbf{T}}(t)) \\
X^{\mathbf{T}}(0) &= X_0 \\
I^{\mathbf{T}}(0) &= i_0
\end{aligned}
$$

Although every program $l$ induces a corresponding agent function $Agent(l, M)$, the action that follows a given percept is not necessarily the agent's "response" to that percept; because of the delay incurred by deliberation, it may only reflect percepts occurring much earlier in the sequence. Furthermore, it is not possible to map every agent function to an implementation $l \in \mathcal{L}_M$. We can define a subset of the set of agent functions $f$ that are implementable on a given architecture $M$ and language $\mathcal{L}_M$:

$$Feasible(M) = \{f \mid \exists l \in \mathcal{L}_M, \ f = Agent(l, M)\}$$

Feasibility is related to, but clearly distinct from, the notion of computability. Computability refers to the existence of a program that *eventually* returns the output specified by a function, whereas feasibility refers to the production of the output at the appropriate point in time. The set of feasible agent functions is therefore much smaller than the set of computable agent functions.

### 3.4 Performance measures for agents

To evaluate an agent's performance in the world, we define a real-valued utility function $U$ on state histories:

$$U : \mathbf{X^T} \to \Re$$

The utility function should be seen as external to the agent and its environment. It defines the problem to be solved by the designer of the agent. *Some* agent designs may incorporate an explicit representation of the utility function, but this is by no means required. We will use the term *task environment* to denote the combination of an environment and a utility function.

Recall that the agent's actions drive the environment $E$ through a particular sequence of states in accordance with the function *effects*$(f, E)$. We can define the value of an agent function $f$ in the environment $E$ as the utility of the state history it generates:

$$V(f, E) = U(\textit{effects}(f, E))$$

If the designer has a set $\mathbf{E}$ of environments with a probability distribution $p$ over them, instead of a single environment $E$, then the value of the agent in $\mathbf{E}$ is defined as the expected value over the elements of $\mathbf{E}$. By a slight abuse of notation,

$$V(f, \mathbf{E}) = \sum_{E \in \mathbf{E}} p(E) V(f, E)$$

We can assign a value $V(l, M, E)$ to a program $l$ executed by the architecture $M$ in the environment $E$ simply by looking at the effect of the agent function implemented by the program:

$$V(l, M, E) = V(\textit{Agent}(l, M), E) = U(\textit{effects}(\textit{Agent}(l, M), E))$$

As above, we can extend this to a set of possible environments as follows:

$$V(l, M, \mathbf{E}) = \sum_{E \in \mathbf{E}} p(E) V(l, M, E)$$

### 3.5 Perfect rationality and bounded optimality

As discussed in Section 2, a perfectly rational agent selects the action that maximizes its expected utility, given the percepts so far. In our framework, this amounts to an agent function that maximizes $V(f, \mathbf{E})$ over all possible agent functions.

**Definition 5** *A perfectly rational agent for a set $\mathbf{E}$ of environments has an agent function $f_{\mathbf{opt}}$ such that*

$$f_{\mathbf{opt}} = \text{argmax}_f(V(f, \mathbf{E}))$$

This definition is a persuasive specification of an optimal agent function for a given set of environments, and underlies several recent projects in intelligent agent design (Dean & Wellman, 1991; Doyle, 1988; Hansson & Mayer, 1989). A direct implementation of this specification, which ignores the delay incurred by deliberation, does not yield a reasonable solution to our problem – the calculation of expected utilities takes time for any real agent. In terms of our simple formal description of agents introduced above, it is easy to see where the difficulty has arisen. In designing the agent program, logicists and decision theorists have concentrated on specifying an optimal agent function $f_{opt}$ in order to guarantee the selection of the best action history. The function $f_{opt}$ is independent of the architecture $M$. Unfortunately, no real program in $\mathcal{L}_M$ implements this function in a non-trivial environment, because optimal actions cannot usually be computed before the next percept arrives. That is, quite frequently, $f_{opt} \notin Feasible(M)$.

Suppose the environment consists of games of chess under tournament rules against some population of human grandmasters, and suppose $M$ is some standard personal computer. Then $f_{opt}$ describes an agent that always plays in such a way as to maximize its total expected points against the opposition, where the maximization is over the *moves* it makes. We claim that no possible program can play this way. It is quite possible, using depth-first alpha-beta search to termination, to execute the program that chooses (say) the optimal minimax move in each situation, but the agent function induced by this program is not the same as $f_{opt}$. In particular, it ignores such percepts as the dropping of its flag indicating a loss on time.

The trouble with the perfect rationality definition arose because of unconstrained optimization over the space of $f$'s in the determination of $f_{opt}$, without regard to feasibility. (Similarly, metalevel rationality assumes unconstrained optimization over the space of deliberations.) To escape this quandary, we propose a machine-dependent standard of rationality, in which we maximize $V$ over the implementable set of agent functions $Feasible(M)$. That is, we impose optimality constraints on *programs* rather than on *agent functions* or *deliberations*.

**Definition 6** *A bounded-optimal agent with architecture $M$ for a set $\mathbf{E}$ of environments has an agent program $l_{opt}$ such that*

$$l_{opt} = \mathrm{argmax}_{l \in \mathcal{L}_M} V(l, M, \mathbf{E})$$

We can see immediately that this specification avoids the most obvious problems with Type I and Type II rationality. Consider our chess example, and suppose the computer has a total program memory of 8 megabytes. Then there are $2^{2^{26}}$ possible programs that can be represented in the machine, of which a much smaller number play legal chess. Under tournament conditions, one or more of these programs will have the best expected performance. Each is a suitable candidate for $l_{opt}$. Thus bounded optimality is, by definition, a feasible specification; moreover, a program that achieves it is highly desirable. We are not yet ready to announce the identity of $l_{opt}$ for chess on an eight-megabyte PC, so we will begin with a more restricted problem.

## 4. Provably Bounded-Optimal Agents

In order to construct a provably bounded optimal agent, we must carry out the following steps:

- Specify the properties of the environment in which actions will be taken, and the utility function on the behaviours.

- Specify a class of machines on which programs are to be run.

- Propose a construction method.

- Prove that the construction method succeeds in building bounded optimal agents.

The methodology is similar to the formal analysis used in the field of optimal control, which studies the design of *controllers* (agents) for *plants* (environments). In optimal control theory, a controller is viewed as an essentially instantaneous implementation of an optimal agent function. In contrast, we focus on the computation time required by the agent, and the relation between computation time and the dynamics of the environment.

## 4.1 Episodic, real-time task environments

In this section, we will consider a restricted class of task environments which we call *episodic* environments. In an episodic task environment, the state history generated by the actions of the agent can be considered as divided into a series of *episodes*, each of which is terminated by an action. Let $\mathbf{A}_\perp \subseteq \mathbf{A}$ be a distinguished set of actions that terminate an episode. The utility of the complete history is given by the sum of the utilities of each episode, which is determined in turn by the state sequence. After each $A \in \mathbf{A}_\perp$, the environment "resets" to a state chosen at random from a stationary probability distribution $P_{init}$. In order to include the effects of the choice of $A$ in the utility of the episode, we notionally divide the environment state into a "configuration" part and a "value" part, such that the configuration part determines the state transitions while the value part determines the utility of a state sequence. Actions in $\mathbf{A}_\perp$ reset the configuration part, while their "value" is recorded in the value part. These restrictions mean that each episode can be treated as a separate decision problem, and translate into the following property: if agent program $l_1$ has higher expected utility on individual episodes than agent $l_2$, it will have higher expected utility in the corresponding episodic task environment.

A real-time task environment is one in which the utility of an action depends on the time at which it is executed. Usually, this dependence will be sufficiently strong to make calculative rationality an unacceptably bad approximation to perfect rationality.

An automated mail sorter[4] provides an illustrative example of an episodic task environment (see Figure 1). Such a machine scans handwritten or printed addresses (zipcodes) on mail pieces and dispatches them to appropriate bins. Each episode starts with the arrival of a new mail piece and terminates with the execution of the physical action recommended by the sorter: routing of the piece to a specific bin. The "configuration part" of the environment corresponds to the letter feeder side, which provides a new, randomly selected letter after the previous letter is sorted. The "value part" of the state corresponds to the state of the receiving bins, which determines the utility of the process. The aim is to maximize the accuracy of sorting while minimizing the reject percentage and avoiding jams. A jam occurs if the current piece is not routed to the appropriate bin, or rejected, before the arrival of the next piece.

We now provide formal definitions for three varieties of real-time task environments: fixed deadlines, fixed time cost and stochastic deadlines.

---

4. See (Sackinger et al. 1992; Boser et al. 1992) for details of an actual system. The application was suggested to us by Bernhard Boser after an early presentation of our work at the 1992 NEC Symposium.

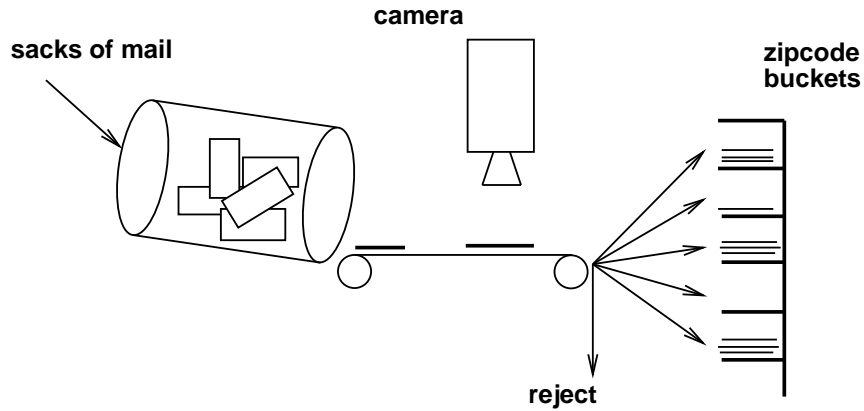

Figure 1: An automated mail-sorting facility provides a simple example of an episodic, real-time task environment.

### 4.1.1 FIXED DEADLINES

The simplest and most commonly studied kind of real-time task environment contains a deadline at a known time. In most work on real-time systems, such deadlines are described informally and systems are built to meet the deadline. Here, we need a formal specification in order to connect the description of the deadline to the properties of agents running in deadline task environments. One might think that deadlines are part of the environment description, but in fact they are mainly realized as constraints on the utility function. One can see this by considering the opposite of a deadline — the "starter's pistol." The two are distinguished by differing constraints on the utilities of acting before or after a specific time.

**Definition 7** *Fixed deadline: The task environment $\langle E, U \rangle$ has a fixed deadline at time $t_d$ if the following conditions hold.*

- *Taking an action in $\mathbf{A}_\perp$ at any time before the deadline results in the same utility:*

$$U([E, A_1^t]) = U([E, A_2^{(t_d-1)} \cdot A_1^{\mathbf{T}}(t)])$$

*where "." denotes sequence concatenation, $t \leq t_d$, $A_1^{\mathbf{T}}(t) \in \mathbf{A}_\perp$, and $A_1^{(t-1)}$ and $A_2^{(t_d-1)}$ contain no action in $\mathbf{A}_\perp$.*

- *Actions taken after $t_d$ have no effect on utility:*

$$U([E, A_1^t]) \leq U([E, A_2^t]) \text{ if } U([E, A_1^{t_d}]) \leq U([E, A_2^{t_d}]) \text{ and } t \geq t_d$$

### 4.1.2 FIXED TIME COST

Task environments with approximately fixed time cost are also very common. Examples include consultations with lawyers, keeping a taxi waiting, or dithering over where to invest one's money. We can define a task environment with fixed time cost $c$ by comparing the utilities of actions taken at different times.

**Definition 8** *Fixed time cost: The task environment $\langle E, U \rangle$ has a fixed time cost if, for any action history prefixes $A_1^{t_1}$ and $A_2^{t_2}$ satisfying*

(1)    $A_1^{\mathbf{T}}(t_1) \in \mathbf{A}_\perp$ *and* $A_2^{\mathbf{T}}(t_2) = A_1^{\mathbf{T}}(t_1)$

(2)    $A_1^{(t_1-1)}$ *and* $A_2^{(t_2-1)}$ *contain no action in* $\mathbf{A}_\perp$

*the utilities differ by the difference in time cost:*

$$U([E, A_2^{t_2}]) = U([E, A_1^{t_1}]) - c(t_2 - t_1)$$

Strictly speaking, there are no task environments with fixed time cost. Utility values have a finite range, so one cannot continue incurring time costs indefinitely. For reasonably short times and reasonably small costs, a linear utility penalty is a useful approximation.

### 4.1.3 Stochastic deadlines

While fixed-deadline and fixed-cost task environments occur frequently in the design of real-time systems, uncertainty about the time-dependence of the utility function is more common. It also turns out to be more interesting, as we see below.

A stochastic deadline is represented by uncertainty concerning the time of occurrence of a fixed deadline. In other words, the agent has a probability distribution $p_d$ for the deadline time $t_d$. We assume that the deadline must come eventually, so that $\sum_{t \in \mathbf{T}} p_d(t) = 1$. We also define the cumulative deadline distribution $P_d$.

If the deadline does not occur at a known time, then we need to distinguish between two cases:

- The agent receives a percept, called a *herald* (Dean & Boddy, 1988), which announces an impending deadline. We model this using a distinguished percept $O_d$:

$$O^{\mathbf{T}}(t_d) = O_d$$

  If the agent responds immediately, then it "meets the deadline."

- No such percept is available, in which case the agent is walking blindfolded towards the utility cliff. By deliberating further, the agent risks missing the deadline but may improve its decision quality. An example familiar to most readers is that of deciding whether to publish a paper in its current form, or to embellish it further and risk being "scooped." We do not treat this case in the current paper.

Formally, the stochastic deadline case is similar to the fixed deadline case, except that $t_d$ is drawn from the distribution $p_d$. The utility of executing an action history prefix $A^t$ in $\mathbf{E}$ is the expectation of the utilities of that state history prefix over the possible deadline times.

**Definition 9** *Stochastic deadline: A task environment class $\langle \mathbf{E}, U \rangle$ of fixed-deadline task environments has a stochastic deadline distributed according to $p_d$ if, for any action history prefix $A^t$,*

$$U([\mathbf{E}, A^t]) = \sum_{t' \in \mathbf{T}} p_d(t') U([E_{t'}, A^t])$$

*where $\langle E_{t'}, U \rangle$ is a task environment in $\langle \mathbf{E}, U \rangle$ with a fixed deadline at $t'$.*

The mail sorter example is well described by a stochastic deadline. The time between the arrival of mail pieces at the image processing station is distributed according to a density function $p_d$, which will usually be Poisson.

## 4.2 Agent programs and agent architecture

We consider simple agent programs for episodic task environments, constructed from elements of a set $\mathbf{R} = \{r_1, \ldots, r_n\}$ of decision procedures or rules. Each decision procedure recommends (but does not execute) an action $A_i \in \mathbf{A}_\perp$, and an agent program is a fixed sequence of decision procedures. For our purposes, a decision procedure is a black box with two parameters:

- a run time $t_i \geq 0$, which is an integer that represents the time taken by the procedure to compute an action.

- a quality $q_i \geq 0$, which is a real number. This gives the expected reward resulting from executing its action $A_i$ at the start of an episode:

$$q_i = U([E, A_i]) \tag{1}$$

Let $M_{\mathcal{J}}$ denote an agent architecture that executes decision procedures in the language $\mathcal{J}$. Let $t_M$ denote the maximum runtime of the decision procedures that can be accommodated in $M$. For example, if the runtime of a feedforward neural network is proportional to its size, then $t_M$ will be the runtime of the largest neural network that fits in $M$.

The architecture $M$ executes an agent program $s = s_1 \ldots s_m$ by running each decision procedure in turn, providing the same input to each as obtained from the initial percept. When a deadline arrives (at a fixed time $t_d$, or heralded by the percept $O_d$), or when the entire sequence has been completed, the agent selects the action recommended by the highest-quality procedure it has executed:

$$\begin{aligned}
M(s, I^{\mathbf{T}}(t_d), O^{\mathbf{T}}(t_d)) &= \langle i_0, \text{action}(I^{\mathbf{T}}(t_d)) \rangle \\
M(s, I^{\mathbf{T}}(t_s), O^{\mathbf{T}}(t_s)) &= \langle i_0, \text{action}(I^{\mathbf{T}}(t_s)) \rangle \text{ where } t_s = \textstyle\sum_{s_i \in s} t_i \\
M(s, I^{\mathbf{T}}(t), O_d) &= \langle i_0, \text{action}(I^{\mathbf{T}}(t)) \rangle
\end{aligned} \tag{2}$$

where $M$ updates the agent's internal state history $I^{\mathbf{T}}(t)$ such that $\text{action}(I^{\mathbf{T}}(t))$ is the action recommended by a completed decision procedure with the highest quality. When this action is executed, the internal state of the agent is re-initialized to $i_0$. This agent design works in all three of the task environment categories described above.

Next we derive the value $V(s, M, E)$ of an agent program $s$ in environment $E$ running on $M$ for the three real-time regimes and show how to construct bounded optimal agents for these task environments.

## 4.3 Bounded optimality with fixed deadlines

From Equation 2, we know that the agent picks the action in $\mathbf{A}_\perp$ recommended by the decision procedure $r$ with the highest quality that is executed before the deadline $t_d$ arrives.

Let $s_1 \ldots s_j$ be the longest prefix of the program $s$ such that $\sum_{i=1}^{j} t_i \leq t_d$. From Definition 7 and Equation 1, it follows that

$$V(s, M, E) = Q_j \tag{3}$$

where $Q_i = \max\{q_1, \ldots, q_i\}$. Given this expression for the value of the agent program, we can easily show the following:

**Theorem 1** *Let $r* = \arg\max_{r_i \in \mathbf{R}, t_i \leq t_d} q_i$. The singleton sequence $r*$ is a bounded optimal program for $M$ in an episodic task environment with a known deadline $t_d$.*

That is, the best program is the single decision procedure of maximum quality whose runtime is less than the deadline.

### 4.4 Bounded optimality with fixed time cost

From Equation 2, we know that the agent picks the action in $\mathbf{A}_\perp$ recommended by the best decision procedure in the sequence, since $M$ runs the entire sequence $s = s_1 \ldots s_m$ when there is no deadline. From Definition 8 and Equation 1, we have

$$V(s, M, E) = Q_m - c \sum_{i=1}^{m} t_i \tag{4}$$

Given this expression for the value of the agent program, we can easily show the following:

**Theorem 2** *Let $r* = \arg\max_{r_i \in \mathbf{R}} q_i - ct_i$. The singleton sequence $r*$ is a bounded optimal program for $M$ in an episodic task environment with a fixed time cost $c$.*

That is, the optimal program is the single decision procedure whose quality, net of time cost, is highest.

### 4.5 Bounded optimality with stochastic deadlines

With a stochastic deadline distributed according to $p_d$, the value of an agent program $s = s_1 \ldots s_m$ is an expectation. From Definition 9, we can calculate this as $\sum_{t \in \mathbf{T}} p_d(t) V(s, M, E_t)$, where $\langle E_t, U \rangle$ is a task environment with a fixed deadline at $t$. After substituting for $V(s, M, E_t)$ from Equation 3, this expression simplifies to a summation, over the procedures in the sequence, of the probability of interruption after the $i^{th}$ procedure in the sequence multiplied by the quality of the best completed decision procedure:

$$V(s) \equiv V(s, M, \mathbf{E}) = \sum_{i=1}^{m} [P_d(\sum_{j=1}^{i+1} t_j) - P_d(\sum_{j=1}^{i} t_j)] Q_i \tag{5}$$

where $P_d(t) = \int_{-\infty}^{t} p_d(t') dt'$ and $P_d(t) = 1$ for $t \geq \sum_{i=1}^{m} t_i$.

A simple example serves to illustrate the value function. Consider $\mathbf{R} = \{r_1, r_2, r_3\}$. The rule $r_1$ has a quality of 0.2 and needs 2 seconds to run: we will represent this by $r_1 = (0.2, 2)$. The other rules are $r_2 = (0.5, 5)$, $r_3 = (0.7, 7)$. The deadline distribution function $p_d$ is a uniform distribution over 0 to 10 seconds. The value of the sequence $r_1 r_2 r_3$ is

$$V(r_1 r_2 r_3) = [.7 - .2].2 + [1 - .7].5 + [1 - 1].7 = .25$$

A geometric intuition is given by the notion of a *performance profile*, as shown in Figure 2.

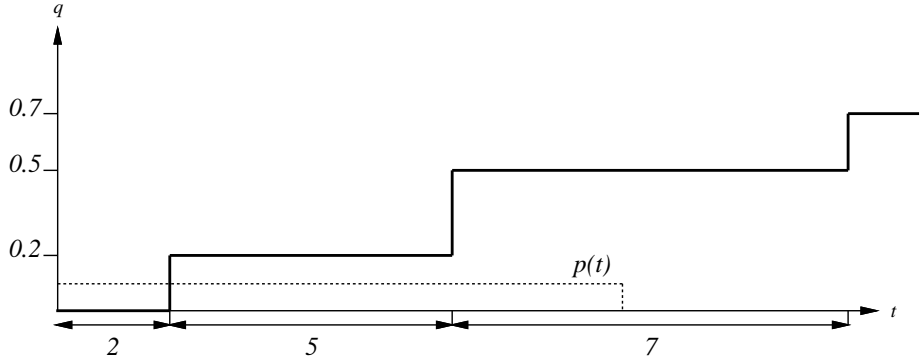

Figure 2: Performance profile for $r_1 r_2 r_3$, with $p_d$ superimposed.

**Definition 10** *Performance profile: For a sequence $s$, the performance profile $Q_s(t)$ gives the quality of the action returned if the agent is interrupted at $t$:*

$$Q_s(t) = max\{q_i : \sum_{j=1}^{i} t_j \le t\}$$

For a uniform deadline density function, the value of a sequence is proportional to the area under the performance profile up to the last possible interrupt time. Note that the height of the profile during the interval of length $t_i$ while rule $i$ is running is the quality of the best of the *previous* rules.

From Definition 10, we have the following obvious property:

**Lemma 1** *The performance profile of any sequence is monotonically nondecreasing.*

It is also the case that a sequence with higher quality decisions at all times is a better sequence:

**Lemma 2** *If $\forall t \ Q_{s_1}(t) \ge Q_{s_2}(t)$, then $V(s_1) \ge V(s_2)$.*

In this case we say that $Q_{s_1}$ *dominates* $Q_{s_2}$.

We can use the idea of performance profiles to establish some useful properties of optimal sequences.

**Lemma 3** *There exists an optimal sequence that is sorted in increasing order of $q$'s.*

Without Lemma 3, there are $\sum_{i=1}^{n} i!$ possible sequences to consider. The ordering constraint eliminates all but $2^n$ sequences. It also means that in proofs of properties of sequences, we now need consider only ordered sequences. In addition, we can replace $Q_i$ in Equation 5 by $q_i$.

The following lemma establishes that a sequence can always be improved by the addition of a better rule at the end:

**Lemma 4** *For every sequence $s = s_1 \ldots s_m$ sorted in increasing order of quality, and single step $z$ with $q_z \ge q_{s_m}$, $V(sz) \ge V(s)$.*

**Corollary 1** *There exists an optimal sequence ending with the highest-quality rule in* **R***.*

The following lemma reflects the obvious intuition that if one can get a better result in less time, there's no point spending more time to get a worse result:

**Lemma 5** *There exists an optimal sequence whose rules are in nondecreasing order of $t_i$.*

We now apply these preparatory results to derive algorithms that construct bounded optimal programs for various deadline distributions.

### 4.5.1 GENERAL DISTRIBUTIONS

For a general deadline distribution, the *dynamic programming* method can be used to obtain an optimal sequence of decision rules in pseudo-polynomial time. We construct an optimal sequence by using the definition of $V(s, M, E)$ in Equation 5. Optimal sequences generated by the methods are ordered by $q_i$, in accordance with Lemma 3.

We construct the table $S(i, t)$, where each entry in the table is the highest value of any sequence that ends with rule $r_i$ at time $t$. We assume the rule indices are arranged in increasing order of quality, and $t$ ranges from the start time 0 to the end time $L = \sum_{r_i \in \mathbf{R}} t_i$. The update rule is:

$$S(i, t) = max_{k \in [0 \ldots i-1]}[S(k, t - t_i) + (q_i - q_k)[1 - P_d(t)]]$$

with boundary condition

$$S(i, 0) = 0 \text{ for each rule } i \text{ and } S(0, t) = 0 \text{ for each time t}$$

From Corollary 1, we can read off the best sequence from the highest value in row $n$ of the matrix $S$.

**Theorem 3** *The DP algorithm computes an optimal sequence in time $O(n^2 L)$ where $n$ is the number of decision procedures in* **R***.*

The dependence on $L$ in the time complexity of the DP algorithm means that the algorithm is not polynomial in the input size. Using standard rounding and scaling methods, however, a fully polynomial approximation scheme can be constructed. Although we do not have a hardness proof for the problem, John Binder (1994) has shown that if the deadline distribution is used as a constant-time oracle for finding values of $P(t)$, any algorithm will require an exponential number of calls to the oracle in the worst case.

### 4.5.2 LONG UNIFORM DISTRIBUTIONS

If the deadline is uniformly distributed over a time interval greater than the sum of the running times of the rules, we will call the distribution a *long uniform* distribution. Consider the rule sequence $s = s_1 \ldots s_m$ drawn from the rule set **R**. With a long uniform distribution, the probability that the deadline arrives during rule $s_i$ of the sequence $s$ is independent of the time at which $s_i$ starts. This permits a simpler form of Equation 5:

$$V(s, M, \mathbf{E}) = \sum_{i=1}^{m-1} P_d(t_{i+1})q_i + q_m(1 - \sum_{i=1}^{m} P_d(t_i)) \tag{6}$$

To derive an optimal sequence under a long uniform distribution, we obtain a recursive specification of the value of a sequence $as$ with $a \in \mathbf{R}$ and $s = s_1 \ldots s_m$ being some sequence in $\mathbf{R}$.

$$V(as, M, \mathbf{E}) = V(s, M, \mathbf{E}) + q_a P_d(t_1) - q_m P_d(t_a) \tag{7}$$

This allows us to define a dynamic programming scheme for calculating an optimal sequence using a state function $S(i, j)$ denoting the highest value of a rule sequence that starts with rule $i$ and ends in rule $j$. From Lemma 3 and Equation 7, the update rule is:

$$S(i, j) = max_{i < k \leq j}[S(k, j) + P_d(t_k)q_i - P_d(t_i)q_j] \tag{8}$$

with boundary condition

$$S(i, i) = (1 - P_d(t_i))q_i \tag{9}$$

From Corollary 1, we know that an optimal sequence for the long uniform distribution ends in $r_n$, the rule with the highest quality in $\mathbf{R}$. Thus, we only need to examine $S(i, n), 1 \leq i \leq n$. Each entry requires $O(n)$ computation, and there are $n$ entries to compute. Thus, the optimal sequence for the long uniform case can be calculated in $O(n^2)$.

**Theorem 4** *An optimal sequence of decision procedures for a long uniform deadline distribution can be determined in $O(n^2)$ time where $n$ is the number of decision procedures in* $\mathbf{R}$.

### 4.5.3 SHORT UNIFORM DISTRIBUTIONS

When $\sum_{i=1}^{n} P_d(t_i) > 1$, for a uniform deadline distribution $P_d$, we call it *short*. This means that some sequences are longer than the last possible deadline time, and therefore some rules in those sequences have no possibility of executing before the deadline. For such sequences, we cannot use Equation 7 to calculate $V(s)$. However, any such sequence can be truncated by removing all rules that would complete execution after the last possible deadline. The value of the sequence is unaffected by truncation, and for truncated sequences the use of Equation 7 is justified. Furthermore, there is an optimal sequence that is a truncated sequence.

Since the update rule 8 correctly computes $S(i, j)$ for truncated sequences, we can use it with short uniform distributions provided we add a check to ensure that the sequences considered are truncated. Unlike the long uniform case, however, the identity of the last rule in an optimal sequence is unknown, so we need to compute all $n^2$ entries in the $S(i, j)$ table. Each entry computation takes $O(n)$ time, thus the time to compute an optimal sequence is $O(n^3)$.

**Theorem 5** *An optimal sequence of decision procedures for a short uniform deadline distribution can be determined in $O(n^3)$ time where $n$ is the number of decision procedures in* $R$.

### 4.5.4 Exponential distributions

For an exponential distribution, $P_d(t) = 1 - e^{-\beta t}$. Exponential distributions allow an optimal sequence to be computed in polynomial time. Let $p_i$ stand for the probability that rule $i$ is interrupted, assuming it starts at 0. Then $p_i = P_d(t_i) = 1 - e^{-\beta t_i}$. For the exponential distribution, $V(s, M, \mathbf{E})$ simplifies out as:

$$V(s, M, \mathbf{E}) = \sum_{i=1}^{m-1} \left[ \Pi_{j=1}^i (1 - p_j) \right] p_{i+1} q_i + \left[ \Pi_{j=1}^m (1 - p_j) \right] q_m$$

This yields a simple recursive specification of the value $V(as, M, \mathbf{E})$ of a sequence that begins with the rule $a$:

$$V(as, M, \mathbf{E}) = (1 - p_a) p_1 q_a + (1 - p_a) V(s, M, \mathbf{E})$$

We will use the state function $S(i, j)$ which represents the highest value of any rule sequence starting with $i$ and ending in $j$.

$$S(i, j) = max_{i < k \leq j} \left[ (1 - p_i) p_k q_i + (1 - p_i) S(k, j) \right]$$

with boundary condition $S(i, i) = q_i (1 - p_i)$. For any given $j$, $S(i, j)$ can be calculated in $O(n^2)$. From Corollary 1, we know that there is an optimal sequence whose last element is the highest-valued rule in $\mathbf{R}$.

**Theorem 6** *An optimal sequence of decision procedures for an exponentially distributed stochastic deadline can be determined in $O(n^2)$ time where $n$ is the number of decision procedures in $\mathbf{R}$.*

The proof is similar to the long uniform distribution case.

## 4.6 Simulation results for a mail-sorter

The preceding results provide a set of algorithms for optimizing the construction of an agent program for a variety of general task environment classes. In this section, we illustrate these results and the possible gains that can be realized in a specific task environment, namely, a simulated mail-sorter.

First, let us be more precise about the utility function $U$ on episodes. There are four possible outcomes; the utility of outcome $i$ is $u_i$.

1. The zipcode is successfully read and the letter is sent to the correct bin for delivery.

2. The zipcode is misread and the letter goes to the wrong bin.

3. The letter is sent to the reject bin.

4. The next letter arrives before the recognizer has finished, and there is a jam. Since letter arrival is heralded, jams cannot occur with the machine architecture given in Equation 2.

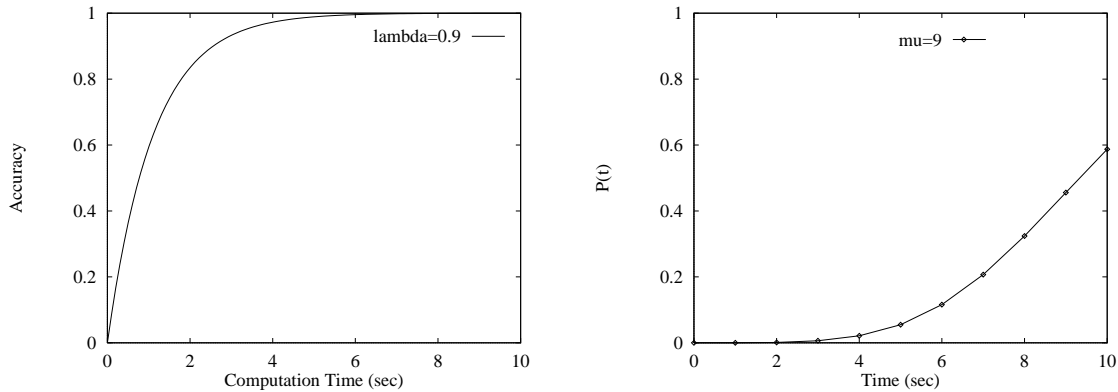

Figure 3: (a) Accuracy profile $(1 - e^{-\lambda x})$, for $\lambda = 0.9$. (b) Poisson arrival distribution, for mean $\mu = 9$ sec

Without loss of generality, we set $u_1 = 1.0$ and $u_2 = 0.0$. If the probability of a rule recommending a correct destination bin is $p_i$, then $q_i = p_i u_1 + (1 - p_i) u_2 = p_i$. We assume that $u_2 \leq u_3$, hence there is a threshold probability below which the letter should be sent to the reject bin instead. We will therefore include in the rule set $\mathbf{R}$ a rule $r_{reject}$ that has zero runtime and recommends rejection. The sequence construction algorithm will then automatically exclude rules with quality lower than $q_{reject} = u_3$. The overall utility for an episode is chosen to be a linear combination of the quality of sorting ($q_i$), the probability of rejection or the rejection rate (given by $P(t_1)$, where $t_1$ is the runtime of the first non-reject rule executed), and the speed of sorting (measured by the arrival time mean).

The agent program in (Boser et al. 1992) uses a single neural network on a chip. We show that under a variety of conditions an optimized sequence of networks can do significantly better than any single network in terms of throughput or accuracy. We examine the following experimental conditions:

- We assume that a network that executes in time $t$ has a recognition accuracy $p$ that depends on $t$. We consider $p = 1 - e^{-\lambda t}$. The particular choice of $\lambda$ is irrelevant because the scale chosen for $t$ is arbitrary. We choose $\lambda = 0.9$, for convenience (Figure 3(a)). We include $r_{reject}$ with $q_{reject} = u_3$ and $t_{reject} = 0$.

- We consider arrival time distributions that are Poisson with varying means. Figure 3(b) shows three example distributions, for means 1, 5, and 9 seconds.

- We create optimized sequences from sets of 40 networks with execution times taken at equal intervals from $t = 1$ to 40.

- We compare

  (a) BO sequence: a bounded optimal sequence;

  (b) Best singleton: the best single rule;

  (c) 50% rule: the rule whose execution time is the mean of the distribution (i.e., it will complete in 50% of cases);

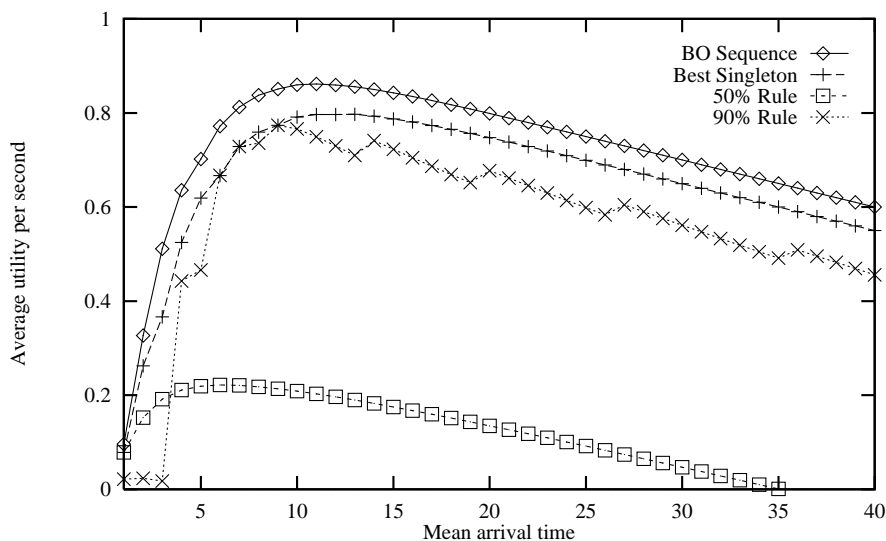

Figure 4: Graph showing the achievable utility per second as a function of the average time per letter, for the four program types. $\lambda = 0.9$.

(d) 90% rule: the rule whose execution time guarantees that it will complete in 90% of cases.

In the last three cases, we add $r_{reject}$ as an initial step; the BO sequence will include it automatically.

- We measure the utility per second as a function of the mean arrival rate (Figure 4). This shows that there is an optimal setting of the sorting machinery at 6 letters per minute (inter-arrival time = 10 seconds) for the bounded optimal program, given that we have fixed $\lambda$ at 0.9.

- Finally, we investigate the effect of the variance of the arrival time on the relative performance of the four program types. For this purpose, we use a uniform distribution centered around 20 seconds but with different widths to vary the variance without affecting the mean (Figure 5).

We notice several interesting things about these results:

- The policy of choosing a rule with a 90% probability of completion performs poorly for rapid arrival rates ($\mu \leq 3$), but catches up with the performance of the best single rule for slower arrival rates ($\mu > 4$). This is an artifact of the exponential accuracy profile for any $\lambda > 0.5$, where the difference in quality of the rules with run times greater than 6 seconds is quite small.

- The policy of choosing a rule with a 50% probability of completion fares as well as the best single rule for very high arrival rates ($\mu \leq 2$), but rapidly diverges from it thereafter, performing far worse for arrival time means greater than 5 seconds.

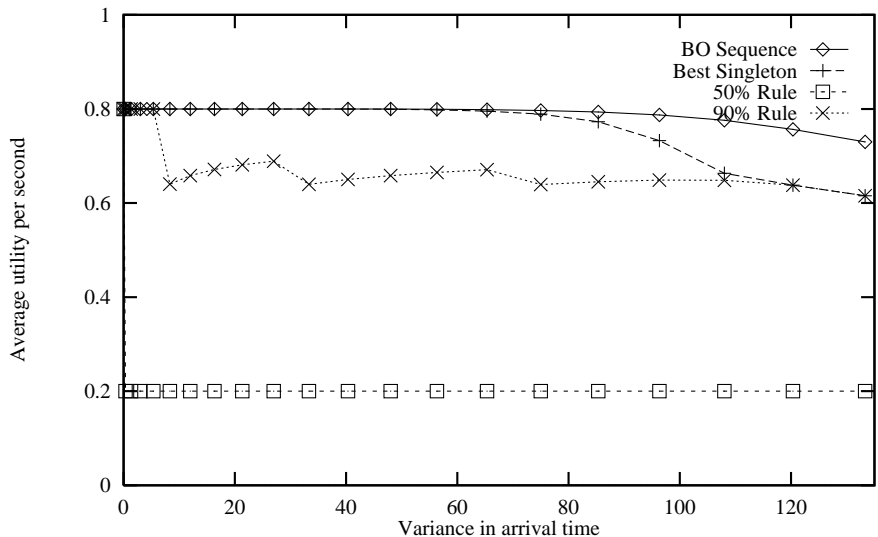

Figure 5: Graphs showing the utility gain per second as a function of the arrival time variance, for the four program types for the uniform distribution with a mean of 20 seconds.

- Both the best sequence and the best single rule give their best overall performance at an arrival rate of around 6 letters per minute. The performance advantage of the optimal sequence over the best single rule is about 7% at this arrival rate. It should be noted that this is a significant performance advantage that is obtainable with no extra computational resources. For slower arrival rates ($\mu \geq 7$), the difference between the performance of the best rule and the best sequence arises from the decreased rejection rate of the best sequence. With the exponential accuracy profile ($\lambda \geq 0.5$) the advantage of running a rule with a shorter completion time ahead of a longer rule is the ability to reduce the probability of rejecting a letter. For high arrival rates (inter-arrival times of 1 to 4 seconds), it is useful to have a few short rules instead of a longer single rule.

- Figure 5 shows that the best sequence performs better than the best single rule as the variance of the arrival time increases.[5] The performance of the optimal sequence also appears to be largely unaffected by variance. This is exactly the behaviour we expect to observe — the ability to run a sequence of rules instead of committing to a single one gives it robustness in the face of increasing variance. Since realistic environments can involve unexpected demands of many kinds, the possession of a variety of default behaviours of graded sophistication would seem to be an optimal design choice for a bounded agent.

---

5. The performance of the 50% rule is flat because the uniform distributions used in this experiment have fixed mean and are symmetric, so that the 50% rule is always the rule that runs for 20 seconds. The 90% rule changes with the variance, and the curve exhibits some discretization effects. These could be eliminated using a finer-grained set of rules.

## 5. Learning Approximately Bounded-Optimal Programs

The above derivations assume that a suitable rule set $\mathbf{R}$ is available *ab initio*, with correct qualities $q_i$ and runtimes $t_i$, and that the deadline distribution is known. In this section, we study ways in which some of this information can be learned, and the implications of this for the bounded optimality of the resulting system. We will concentrate on learning rules and their qualities, leaving runtimes and deadline distributions for future work.

The basic idea is that the learning algorithms will converge, over time, to a set of optimal components — the most accurate rules and the most accurate quality estimates for them. As this happens, the value of the agent constructed from the rules, using the quality estimates, converges to the value of $l_{\mathrm{opt}}$. Thus there are two sources of suboptimality in the learned agent:

- The rules in $\mathbf{R}$ may not be the best possible rules — they may recommend actions that are of lower utility than those that would be recommended by some other rules.

- There may be errors in estimating the expected utility of the rule. This can cause the algorithms given above to construct suboptimal sequences, even if the best rules are available.

Our notional method for constructing bounded optimal agents (1) learns sets of individual decision procedures from episodic interactions, and (2) arranges them in a sequence using one of the algorithms described earlier so that the performance of an agent using the sequence is at least as good as that of any other such agent. We assume a parameterized learning algorithm $L_{\mathcal{J},k}$ that will be used to learn one rule for each possible runtime $k \in \{1, \ldots, t_M\}$. Since there is never a need to include two rules with the same runtime in the $\mathbf{R}$, this obviates the need to consider the entire rule language $\mathcal{J}$ in the optimization process.

Our setting places somewhat unusual requirements on the learning algorithm. Like most learning algorithms, $L_{\mathcal{J},k}$ works by observing a collection $T$ of training episodes in $\mathbf{E}$, including the utility obtained for each episode. We do not, however, make any assumptions about the form of the *correct* decision rule. Instead, we make assumptions about the hypotheses, namely that they come from some finite language $\mathcal{J}_k$, the set of programs in $\mathcal{J}$ of complexity at most $k$. This setting has been called the *agnostic learning* setting by Kearns, Schapire and Sellie (1992), because no assumptions are made about the environment at all. It has been shown (Theorems 4 and 5 in Kearns, Schapire and Sellie, 1992) that, for some languages $\mathcal{J}$, the error in the learned approximation can be bounded to within $\epsilon$ of the best rule in $\mathcal{J}_k$ that fits the examples, with probability $1 - \delta$. The sample size needed to guarantee these bounds is polynomial in the complexity parameter $k$, as well as $\frac{1}{\epsilon}$ and $\frac{1}{\delta}$.

In addition to constructing the decision procedures, $L_{\mathcal{J},k}$ outputs estimates of their quality $q_i$. Standard Chernoff-Hoeffding bounds can be used to limit the error in the quality estimate to be within $\epsilon_q$ with probability $1 - \delta_q$. The sample size for the estimation of quality is also polynomial in $\frac{1}{\epsilon_q}$ and $\frac{1}{\delta_q}$.

Thus the error in each agnostically learned rule is bounded to within $\epsilon$ of the best rule in its complexity class with probability $1 - \delta$. The error in the quality estimation of these rules is bounded by $\epsilon_q$ with probability $1 - \delta_q$. From these bounds, we can calculate a bound on the utility deficit in the agent program that we construct, in comparison to $l_{\mathrm{opt}}$:

**Theorem 7** *Assume an architecture $M_{\mathcal{J}}$ that executes sequences of decision procedures in an agnostically learnable language $\mathcal{J}$ whose runtimes range over $[1..t_M]$. For real time task environments with fixed time cost, fixed deadline, and stochastic deadline, we can construct a program $l$ such that*

$$V(l_{\text{opt}}, M, \mathbf{E}) - V(l, M, \mathbf{E}) \leq \epsilon + 2\epsilon_q$$

*with probability greater than $1 - m(\delta + \delta_q)$, where $m$ is the number of decision procedures in $l_{\text{opt}}$.*

**Proof:** We prove this theorem for the stochastic deadline regime, where the bounded optimal program is a sequence of decision procedures. The proofs for the fixed cost and fixed deadline regimes, where the bounded optimal program is a singleton, follow as a special case. Let the best decision procedures for $\mathbf{E}$ be the set $\mathbf{R}^* = \{r_1^*, \ldots, r_n^*\}$, and let $l_{\text{opt}} = s_1^* \ldots s_m^*$ be an optimal sequence constructed from $\mathbf{R}^*$. Let $\mathbf{R} = \{r_1, \ldots r_n\}$ be the set of decision procedures returned by the learning algorithm. With probability greater than $1 - m\delta$, $q_i^* - q_i \leq \epsilon$ for all $i$, where $q_i$ refers to the *true* quality of $r_i$. The error in the *estimated* quality $\hat{q}_i$ of decision procedure $r_i$ is also bounded: with probability greater than $1 - m\delta_q$, $|\hat{q}_i - q_i| \leq \epsilon_q$ for all $i$.

Let $s = s_1 \ldots s_m$ be those rules in $\mathbf{R}$ that come from the same runtime classes as the rules $s_1^* \ldots s_m^*$ in $\mathbf{R}^*$. Then, by Equation 5, we have

$$V(l_{\text{opt}}, M, \mathbf{E}) - V(s, M, \mathbf{E}) \leq \epsilon$$

because the error in $V$ is a weighted average of the errors in the individual $q_i$. Similarly, we have

$$|\hat{V}(s, M, \mathbf{E}) - V(s, M, \mathbf{E})| \leq \epsilon_q$$

Now suppose that the sequence construction algorithm applied to $\mathbf{R}$ produces a sequence $l = s_1' \ldots s_l'$. By definition, this sequence *appears* to be optimal according to the estimated value function $\hat{V}$. Hence

$$\hat{V}(l, M, \mathbf{E}) \geq \hat{V}(s, M, \mathbf{E})$$

As before, we can bound the error on the estimated value:

$$|\hat{V}(l, M, \mathbf{E}) - V(l, M, \mathbf{E})| \leq \epsilon_q$$

Combining the above inequalities, we have

$$V(l_{\text{opt}}, M, \mathbf{E}) - V(l, M, \mathbf{E}) \leq \epsilon + 2\epsilon_q$$

$\square$

Although the theorem has practical applications, it is mainly intended as an illustration of how a learning procedure can converge on a bounded optimal configuration. With some additional work, more general error bounds can be derived for the case in which the rule execution times $t_i$ and the real-time utility variation (time cost, fixed deadline, or deadline distribution) are all estimated from the training episodes. We can also obtain error bounds for the case in which the rule language $\mathcal{J}$ is divided up into a smaller number of coarser runtime classes, rather than the potentially huge number that we currently use.

## 6. Asymptotic Bounded Optimality

The strict notion of bounded optimality may be a useful philosophical landmark from which to explore artificial intelligence, but it may be too strong to allow many interesting, general results to be obtained. The same observation can be made in ordinary complexity theory: although absolute efficiency is the aim, asymptotic efficiency is the game. That a sorting algorithm is $O(n \log n)$ rather than $O(n^2)$ is considered significant, but replacing a "multiply by 2" by a "shift-left 1 bit" is not considered a real advance. The slack allowed by the definitions of complexity classes is essential in building on earlier results, in obtaining robust results that are not restricted to specific implementations, and in analysing the complexity of algorithms that use other algorithms as subroutines. In this section, we begin by reviewing classical complexity. We then propose definitions of asymptotic bounded optimality that have some of the same advantages, and show that classical optimality is a special case of asymptotic bounded optimality. Lastly, we report on some preliminary investigations into the use of asymptotic bounded optimality as a theoretical tool in constructing universal real-time systems.

### 6.1 Classical complexity

A *problem*, in the classical sense, is defined by a pair of predicates $\phi$ and $\psi$ such that output $z$ is a solution for input $x$ if and only if $\phi(x)$ and $\psi(x, z)$ hold. A *problem instance* is an input satisfying $\phi$, and an algorithm for the problem class always terminates with an output $z$ satisfying $\psi(x, z)$ given an input $x$ satisfying $\phi(x)$. Asymptotic complexity describes the growth rate of the worst-case runtime of an algorithm as a function of the input size. We can define this formally as follows. Let $T_a(x)$ be the runtime of algorithm $a$ on input $x$, and let $T_a^*(n)$ be the maximum runtime of $a$ on any input of size $n$. Then algorithm $a$ has complexity $O(f(n))$ if

$$\exists k, n_0 \; \forall n \; n > n_0 \Rightarrow T_a^*(n) \leq k f(n)$$

Intuitively, a classically optimal algorithm is one that has the lowest possible complexity. For the purposes of constructing an asymptotic notion of bounded optimality, it will be useful to have a definition of classical optimality that does not mention the complexity directly. This can be done as follows:

**Definition 11** *Classically optimal algorithm: An algorithm $a$ is classically optimal if and only if*

$$\exists k, n_0 \; \forall a', n \; n > n_0 \Rightarrow T_a^*(n) \leq k T_{a'}^*(n)$$

To relate classical complexity to our framework, we will need to define the special case of task environments in which traditional programs are appropriate. In such task environments, an input is provided to the program as the initial percept, and the utility function on environment histories obeys the following constraint:

**Definition 12** *Classical task environment: $\langle E_P, U \rangle$ is a classical task environment for problem $P$ if*

$$V(l, M, E_P) \;=\; \begin{cases} u(T(l, M, E_P)) & \textit{if } l \textit{ outputs a correct solution for } P \\ 0 & \textit{otherwise} \end{cases}$$

599

*where $T(l, M, E_P)$ is the running time for $l$ in $E_P$ on $M$, $M$ is a universal Turing machine, and $u$ is some positive decreasing function.*

The notion of a *problem class* in classical complexity theory thus corresponds to a class of classical task environments of unbounded complexity. For example, the Traveling Salesperson Problem contains instances with arbitrarily large numbers of cities.

## 6.2 Varieties of asymptotic bounded optimality

The first thing we will need is a complexity measure on environments. Let $n(E)$ be a suitable measure of the complexity of an environment. We will assume the existence of environment classes that are of unbounded complexity. Then, by analogy with the definition of classical optimality, we can define a worst-case notion of asymptotic bounded optimality (ABO). Letting $V^*(l, M, n, \mathbf{E})$ be the minimum value of $V(l, M, E)$ for all $E$ in $\mathbf{E}$ of complexity $n$, we have

**Definition 13** *Worst-case asymptotic bounded optimality: an agent program $l$ is timewise (or spacewise) worst-case asymptotically bounded optimal in $\mathbf{E}$ on $M$ iff*

$$\exists k, n_0 \; \forall l', n \;\; n > n_0 \Rightarrow V^*(l, kM, n, \mathbf{E}) \geq V^*(l', M, n, \mathbf{E})$$

*where $kM$ denotes a version of the machine $M$ speeded up by a factor $k$ (or with $k$ times more memory).*

In English, this means that the program is basically along the right lines if it just needs a faster (larger) machine to have worst-case behaviour as good as that of any other program in all environments.

If a probability distribution is associated with the environment class $\mathbf{E}$, then we can use the expected value $V(l, M, \mathbf{E})$ to define an average-case notion of ABO:

**Definition 14** *Average-case asymptotic bounded optimality: an agent program $l$ is timewise (or spacewise) average-case asymptotically bounded optimal in $\mathbf{E}$ on $M$ iff*

$$\exists k \; \forall l' \; V(l, kM, \mathbf{E}) \geq V(l', M, \mathbf{E})$$

For both the worst-case and average-case definitions of ABO, we would be happy with a program that was ABO for a nontrivial environment on a nontrivial architecture $M$, unless $k$ were enormous.[6] In the rest of the paper, we will use the worst-case definition of ABO. Almost identical results can be obtained using the average-case definition.

The first observation that can be made about ABO programs is that classically optimal programs are a special case of ABO programs:[7]

---

6. The classical definitions allow for optimality up to a constant factor $k$ in the runtime of the algorithms. One might wonder why we chose to use the constant factor to expand the machine capabilities, rather than to increase the time available to the program. In the context of ordinary complexity theory, the two alternatives are exactly equivalent, but in the context of general time-dependent utilities, only the former is appropriate. It would not be possible to simply "let $l$ run $k$ times longer," because the programs we wish to consider control their own execution time, trading it off against solution quality. One could imagine slowing down the entire environment by a factor of $k$, but this is merely a less realistic version of what we propose.

7. This connection was suggested by Bart Selman.

**Theorem 8** *A program is classically optimal for a given problem P if and only if it is timewise worst-case ABO for the corresponding classical task environment class* $\langle \mathbf{E}_P, U \rangle$.

This observation follows directly from Definitions 11, 12, and 13.

In summary, the notion of ABO will provide the same degree of theoretical robustness and machine-independence for the study of bounded systems as asymptotic complexity does for classical programs. Having set up a basic framework, we can now begin to exercise the definitions.

## 6.3 Universal asymptotic bounded optimality

Asymptotic bounded optimality is defined with respect to a specific value function $V$. In constructing real-time systems, we would prefer a certain degree of independence from the temporal variation in the value function. We can achieve this by defining a family $\mathcal{V}$ of value functions, differing only in their temporal variation. By this we mean that the value function preserves the preference ordering of external actions over time, with all value functions in the family having the same preference ordering.[8]

For example, in the fixed-cost regime we can vary the time cost $c$ to generate a family of value functions; in the stochastic deadline case, we can vary the deadline distribution $P_d$ to generate another family. Also, since each of the three regimes uses the same quality measure for actions, then the union of the three corresponding families is also a family. What we will show is that a single program, which we call a universal program, can be asymptotically bounded-optimal regardless of which value function is chosen within any particular family.

**Definition 15** *Universal asymptotic bounded optimality (UABO): An agent program $l$ is UABO in environment class $\mathbf{E}$ on $M$ for the family of value functions $\mathcal{V}$ iff $l$ is ABO in $\mathbf{E}$ on $M$ for every $V_i \in \mathcal{V}$.*

A UABO program must compete with the ABO programs for every individual value function in the family. A UABO program is therefore a universal real-time solution for a given task. Do UABO programs exist? If so, how can we construct them?

It turns out that we can use the scheduling construction from (Russell & Zilberstein, 1991) to design UABO programs. This construction was designed to reduce task environments with unknown interrupt times to the case of known deadlines, and the same insight applies here. The construction requires the architecture $M$ to provide program concatenation (e.g., the LISP `prog` construct), a conditional-return construct, and the null program $\Lambda$. The universal program $l_U$ has the form of a concatenation of individual programs of increasing runtime, with an appropriate termination test after each. It can be written as

$$l_U = [l_0 \cdot l_1 \cdots l_j \cdots]$$

where each $l_j$ consists of a program and a termination test. The program part in $l_j$ is any program in $\mathcal{L}_M$ that is ABO in $\mathbf{E}$ for a value function $V_j$ that corresponds to a fixed deadline at $t_d = 2^j \epsilon$, where $\epsilon$ is a time increment smaller than the execution time of any non-null program in $\mathcal{L}_M$.

---

8. The value function must therefore be *separable* (Russell & Wefald, 1989), since this preservation of rank order allows a separate time cost to be defined. See chapter 9 of (Keeney & Raiffa, 1976) for a thorough discussion of time-dependent utility.

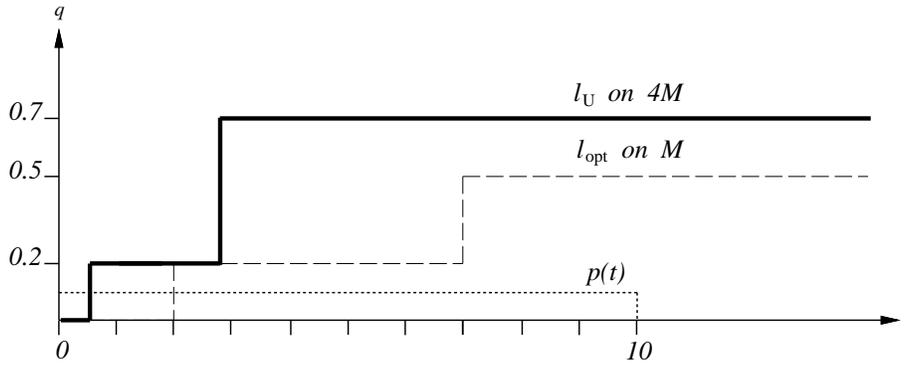

Figure 6: Performance profiles for $l_U$ running on $4M$, and for $l_{opt}$ running on $M$

Before proceeding to a statement that $l_U$ is indeed UABO, let us look at an example. Consider the simple, sequential machine architecture described earlier. Suppose we can select rules from a three-rule set with $r_1 = (0.2, 2)$, $r_2 = (0.5, 5)$ and $r_3 = (0.7, 7)$. Since the shortest runtime of these rules is 2 seconds, we let $\epsilon = 1$. Then we look at the optimal programs $l_0, l_1, l_2, l_3, \ldots$ for the fixed-deadline task environments with $t_d = 1, 2, 4, 8, \ldots$. These are:

$$l_0 = \Lambda; \quad l_1 = r_1; \quad l_2 = r_1; \quad l_3 = r_3; \quad \ldots$$

Hence the sequence of programs in $l_U$ is $[\Lambda, r_1, r_1, r_3, \ldots]$.

Now consider a task environment class with a value function $V_i$ that specifies a stochastic deadline uniformly distributed over the range $[0 \ldots 10]$. For this class, $l_{opt} = r_1 r_2$ is a bounded optimal sequence.[9] It turns out that $l_U$ has higher utility than $l_{opt}$ provided it is run on a machine that is four times faster. We can see this by plotting the two performance profiles: $Q_U$ for $l_U$ on $4M$ and $Q_{opt}$ for $l_{opt}$ on $M$. $Q_U$ dominates $Q_{opt}$, as shown in Figure 6.

To establish that the $l_U$ construction yields UABO programs in general, we need to define a notion of worst-case performance profile. Let $Q^*(t, l, M, n, \mathbf{E})$ be the minimum value obtained by interrupting $l$ at $t$, over all $E$ in $\mathbf{E}$ of complexity $n$. We know that each $l_j$ in $l_U$ satisfies the following:

$$\forall l', n \; n > n_j \Rightarrow V_j^*(l_j, k_j M, n, \mathbf{E}) \geq V_j^*(l', M, n, \mathbf{E})$$

for constants $k_j$, $n_j$. The aim is to prove that

$$\forall V_i \in \mathcal{V} \; \exists k, n_0 \; \forall l', n \; n > n_0 \Rightarrow V_i^*(l_U, kM, n, \mathbf{E}) \geq V_i^*(l', M, n, \mathbf{E})$$

Given the definition of worst-case performance profile, it is fairly easy to show the following lemma (the proof is essentially identical to the proof of Theorem 1 in Russell and Zilberstein, 1991):

---

9. Notice that, in our simple model, the output quality of a rule depends only on its execution time and not on the input complexity. This also means that worst-case and average-case behaviour are the same.

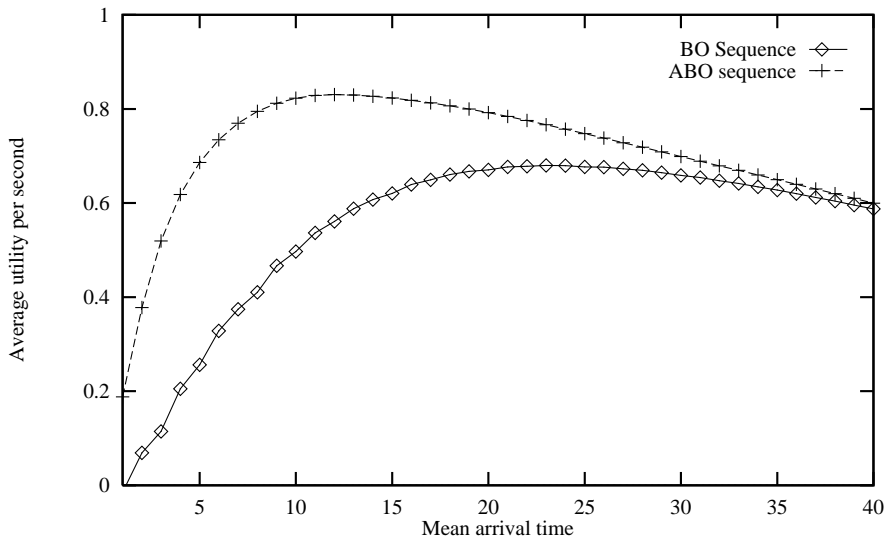

Figure 7: Throughput and accuracy improvement of $l_U$ over $l_{opt}^i$, as a function of mean arrival time, $\lambda = 0.2$, Poisson arrivals.

**Lemma 6** *If $l_U$ is a universal program in $\mathbf{E}$ for $\mathcal{V}$, and $l_i$ is ABO on $M$ in $\mathbf{E}$ for $V_i \in \mathcal{V}$, then $Q^*(t, l_U, kM, n, \mathbf{E})$ dominates $Q^*(t, l_i, M, n, \mathbf{E})$ for $k \geq 4 \max_j k_j$, $n > \max_j n_j$.*

This lemma establishes that, for a small constant penalty, we can ignore the specific real-time nature of the task environment in constructing bounded optimal programs. However, we still need to deal with the issue of termination. It is not possible in general for $l_U$ to terminate at an appropriate time without access to information concerning the time-dependence of the utility function. For example, in a fixed-time-cost task environment, the appropriate termination time depends on the value of the time cost $c$.

For the general case with deterministic time-dependence, we can help out $l_U$ by supplying, for each $V_i$, an "aspiration level" $Q_i^*(t_i, l_i, M, n, \mathbf{E})$, where $t_i$ is the time at which $l_i$ acts. $l_U$ terminates when it has completed an $l_j$ such that $q_j \geq Q_i^*(t_i, l_i, M, n, \mathbf{E})$. By construction, this will happen no later than $t_i$ because of Lemma 6.

**Theorem 9** *In task environments with deterministic time-dependence, an $l_U$ with a suitable aspiration level is UABO in $\mathbf{E}$ on $M$.*

With deadline heralds, the termination test is somewhat simpler and does not require any additional input to $l_U$.

**Theorem 10** *In a task environment with stochastic deadlines, $l_U$ is UABO in $\mathbf{E}$ on $M$ if it terminates when the herald arrives.*

Returning to the mail-sorting example, it is fairly easy to see that $l_U$ (which consists of a sequence of networks, like the optimal programs for the stochastic deadline case) will be ABO in the fixed-deadline regime. It is not so obvious that it is also ABO in any particular stochastic deadline case — recall that both regimes can be considered as a single family. We have programmed a constructor function for universal programs, and applied it to the mail-sorter environment class. Varying the letter arrival distribution gives us different value functions $V_i \in \mathcal{V}$. Figure 7 shows that $l_U$ (on $4M$) has higher throughput and accuracy than $l_{\mathrm{opt}}^i$ across the entire range of arrival distributions.

Given the existence of UABO programs, it is possible to consider the behaviour of compositions thereof. The simplest form of composition is functional composition, in which the output of one program is used as input by another. More complex, nested compositional structures can be entertained, including loops and conditionals (Zilberstein, 1993). The main issue in constructing UABO compositions is how to allocate time among the components. Provided that we can solve the time allocation problem when we know the total runtime allowed, we can use the same construction technique as used above to generate composite UABO programs, where optimality is among all possible compositions of the components. Zilberstein and Russell (1993), show that the allocation problem can be solved in linear time in the size of the composite system, provided the composition is a tree of bounded degree.

## 7. Conclusions And Further Work

We examined three possible formal bases for artificial intelligence, and concluded that bounded optimality provides the most appropriate goal in constructing intelligent systems. We also noted that similar notions have arisen in philosophy and game theory for more or less the same reason: the mismatch between classically optimal actions and what we have called *feasible* behaviours—those that can be generated by an agent program running on a computing device of finite speed and size.

We showed that with careful specification of the task environment and the computing device one can design provably bounded-optimal agents. We exhibited only very simple agents, and it is likely that bounded optimality in the strict sense is a difficult goal to achieve when a larger space of agent programs is considered. More relaxed notions such as asymptotic bounded optimality (ABO) may provide more theoretically robust tools for further progress. In particular, ABO promises to yield useful results on composite agent designs, allowing us to separate the problem of designing complex ABO agents into a discrete structural problem and a continuous temporal optimization problem that is tractable in many cases. Hence, we have reason to be optimistic that artificial intelligence can be usefully characterized as the study of bounded optimality. We may speculate that provided the computing device is neither too small (so that small changes in speed or size cause significant changes in the optimal program design) nor too powerful (so that classically optimal decisions can be computed feasibly), ABO designs should be stable over reasonably wide variations in machine speed and size and in environmental complexity. The *details* of the optimal designs may be rather arcane, and learning processes will play a large part in their discovery; we expect that the focus of this type of research will be more on questions of convergence to optimality for various structural classes than on the end result itself.

Perhaps the most important implication, beyond the conceptual foundations of the field itself, is that research on bounded optimality applies, by design, to the *practice* of artificial intelligence in a way that idealized, infinite-resource models may not. We have given, by way of illustrating this definition, a bounded optimal agent: the design of a simple system consisting of sequences of decision procedures that is provably better than any other program in its class. A theorem that exhibits a bounded optimal design translates, by definition, into an agent whose actual behaviour is desirable.

There appear to be plenty of worthwhile directions in which to continue the exploration of bounded optimality. From a foundational point of view, one of the most interesting questions is how the concept applies to agents that can incorporate a learning component. (Note that in section 5, the learning algorithm was external to the agent.) In such a case, there will not necessarily be a largely stable bounded optimal configuration if the agent program is not large enough; instead, the agent will have to adapt to a shorter-term horizon and rewrite itself as it becomes obsolete.

With results on the preservation of ABO under composition, we can start to examine much more interesting architectures than the simple production system studied above. For example, we can look at optimal search algorithms, where the algorithm is constrained to apply a metalevel decision procedure at each step to decide which node to expand, if any (Russell & Wefald, 1989). We can also extend the work on asymptotic bounded optimality to provide a utility-based analogue to "big-O" notation for describing the performance of agent designs, including those that are suboptimal.

In the context of computational learning theory, it is obvious that the stationarity requirement on the environment, which is necessary to satisfy the preconditions of PAC results, is too restrictive. The fact that the agent learns may have some effect on the distribution of future episodes, and little is known about learning in such cases (Aldous & Vazirani, 1990). We could also relax the deterministic and episodic requirement to allow non-immediate rewards, thereby making connections to current research on reinforcement learning.

The computation scheduling problem we examined is interesting in itself, and does not appear to have been studied in the operations research or combinatorial optimization literature. Scheduling algorithms usually deal with physical rather than computational tasks, hence the objective function usually involves summation of outputs rather than picking the best. We would like to resolve the formal question of its tractability in the general case, and also to look at cases in which the solution qualities of individual processes are interdependent (such as when one can use the results of another). Practical extensions include computation scheduling for parallel machines or multiple agents, and scheduling combinations of computational and physical (e.g., job-shop and flow-shop) processes, where objective functions are a combination of summation and maximization. The latter extension broadens the scope of applications considerably. An industrial process, such as designing and manufacturing a car, consists of both computational steps (design, logistics, factory scheduling, inspection etc.) and physical processes (stamping, assembling, painting etc.). One can easily imagine many other applications in real-time financial, industrial, and military contexts.

It may turn out that bounded optimality is found wanting as a theoretical framework. If this is the case, we hope that it is refuted in an interesting way, so that a better framework can be created in the process.

## Appendix: Additional Proofs

This appendix contains formal proofs for three subsidiary lemmata in the main body of the paper.

**Lemma 3** *There exists an optimal sequence that is sorted in increasing order of $q$'s.*

**Proof:** Suppose this is not the case, and $s$ is an optimal sequence. Then there must be two adjacent rules $i$, $i+1$ where $q_i > q_{i+1}$ (see Figure 8). Removal of rule $i+1$ yields a sequence $s'$ such that $Q_{s'}(t) \geq Q_s(t)$, from Lemma 1 and the fact that $t_{i+2} \leq t_{i+1} + t_{i+2}$. By Lemma 2, $s'$ must also be optimal. We can repeat this removal process until $s'$ is ordered by $q_i$, proving the theorem by reductio ad absurdum.$\Box$

**Lemma 4** *For every sequence $s = s_1 \ldots s_m$ sorted in increasing order of quality, and single step $z$ with $q_z \geq q_{s_m}$, $V(sz) \geq V(s)$.*

**Proof:** We calculate $V(sz) - V(s)$ using Equation 5 and show that it is non-negative:

$$
\begin{aligned}
V(sz) - V(s) &= q_z[1 - P_d((\textstyle\sum_{j=1}^m t_j) + t_z)] - q_m[1 - P_d((\textstyle\sum_{j=1}^m t_j) + t_z)] \\
&= (q_z - q_m)[1 - P_d((\textstyle\sum_{j=1}^m t_j) + t_z)]
\end{aligned}
$$

which is non-negative since $q_z \geq q_m$.$\Box$

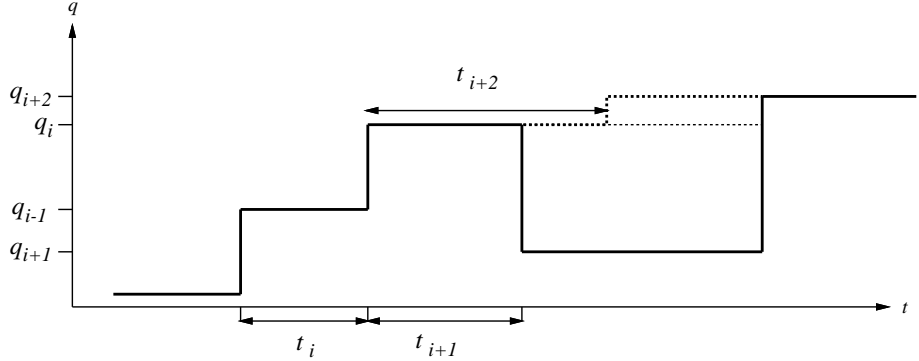

Figure 8: Proof for ordering by $q_i$; lower dotted line indicates original profile; upper dotted line indicates profile after removal of rule $i+1$.

**Lemma 5** *There exists an optimal sequence whose rules are in nondecreasing order of $t_i$.*

**Proof:** Suppose this is not the case, and $s$ is an optimal sequence. Then there must be two adjacent rules $i$, $i+1$ where $q_i \leq q_{i+1}$ and $t_i > t_{i+1}$ (see Figure 9). Removal of rule $i$ yields a sequence $s'$ such that $Q_{s'}(t) \geq Q_s(t)$, from Lemma 1. By Lemma 2, $s'$ must also be optimal. We can repeat this removal process until $s'$ is ordered by $t_i$, proving the theorem by reductio ad absurdum.$\Box$

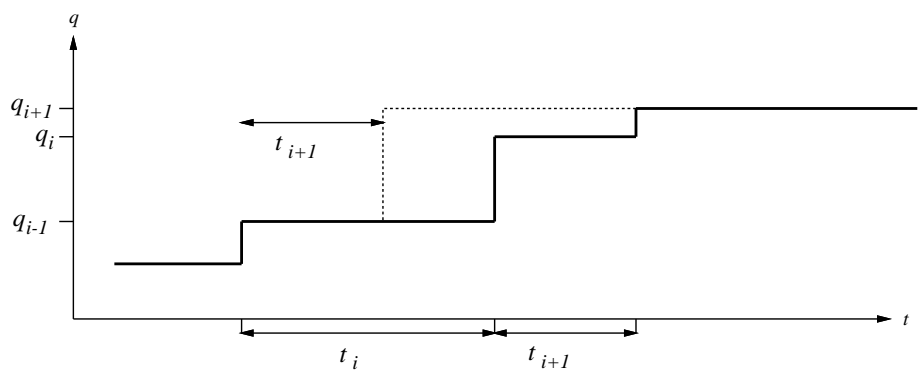

Figure 9: Proof for ordering by $t_i$; dotted line indicates profile after removal of rule $i$.

## Acknowledgements

We would like to acknowledge stimulating discussions with Michael Fehling, Michael Genesereth, Russ Greiner, Eric Horvitz, Henry Kautz, Daphne Koller, and Bart Selman on the subject of bounded optimality; with Dorit Hochbaum, Nimrod Megiddo, and Kevin Glazebrook on the subject of dynamic programming for scheduling problems; and with Nick Littlestone and Michael Kearns on the subject of agnostic learning. We would also like to thank the reviewers for their many constructive suggestions. Many of the early ideas on which this work is based arose in discussions with the late Eric Wefald. Thanks also to Ron Parr for his work on the uniform-distribution case, Rhonda Righter for extending the results to the exponential distribution, and Patrick Zieske for help in implementing the dynamic programming algorithm. The first author was supported by NSF grants IRI-8903146, IRI-9211512 and IRI-9058427, by a visiting fellowship from the SERC while on sabbatical in the UK, and by the NEC Research Institute. The second author was supported by NSF grant IRI-8902721.